# Temporal Extension Module for Skeleton-Based Action Recognition

Yuya Obinata and Takuma Yamamoto
Digital Innovation Core Unit
FUJITSU LABORATORIES LTD.
Kanagawa, Japan
{obinata.yuya, takuma.yamamoto}@fujitsu.com

*Abstract*—We present a module that extends the temporal graph of a graph convolutional network (GCN) for action recognition with a sequence of skeletons. Existing methods attempt to represent a more appropriate spatial graph on an intra-frame, but disregard optimization of the temporal graph on the inter-frame. Concretely, these methods connect between vertices corresponding only to the same joint on the inter-frame. In this work, we focus on adding connections to neighboring multiple vertices on the inter-frame and extracting additional features based on the extended temporal graph. Our module is a simple yet effective method to extract correlated features of multiple joints in human movement. Moreover, our module aids in further performance improvements, along with other GCN methods that optimize only the spatial graph. We conduct extensive experiments on two large datasets, NTU RGB+D and Kinetics-Skeleton, and demonstrate that our module is effective for several existing models and our final model achieves state-of-the-art performance.

## I. INTRODUCTION

Action recognition is a challenging task in computer vision and is on the cusp of applications toward understanding human activity [1] and human social behavior [2]. There are two mainstream methods using RGB images or a sequence of skeletons for action recognition. Action recognition with RGB images may achieve high performance. However, the method needs high-cost computing resources for real-time processing because the method handles several hundred pixels per image to extract features. The method is also affected by noise from various illumination conditions and the background. In contrast, action recognition with a sequence of skeletons, which represent a sequence of 2D or 3D coordinates as human body joints and trajectories, requires lower-cost computing resources than action recognition with RGB images because the method handles only a few dozen joints per skeleton. Moreover, skeletons are robust for the noise described above. Therefore, action recognition with a sequence of skeletons is suitable for practical applications. High-accuracy pose estimation methods such as OpenPose [5] and CPN [6] also support the practicality of the method. Thus, we focus on action recognition with a sequence of skeletons (skeleton-based action recognition) in this work.

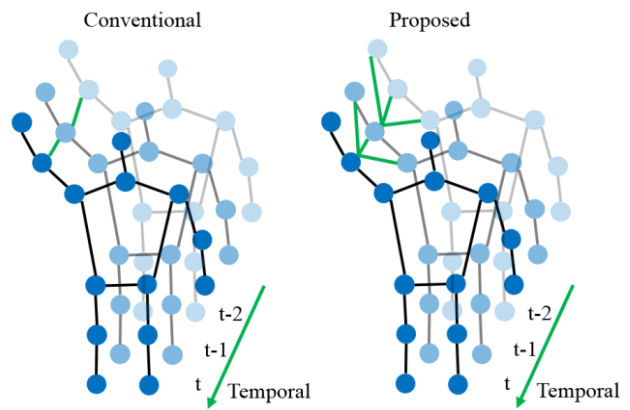

Fig. 1. Our concept. We draw vertices corresponding to human body joints with dots and edges corresponding to the joint connection with the black solid line as the spatial graph. The green solid line represents the edge for the temporal graph. In the conventional temporal graph, vertices corresponding only to the same joint on the inter-frame are connected (left). In this work, we focus on extending the temporal graph and connect neighboring multiple vertices as well as the same vertex on the inter-frame (right).

In skeleton-based action recognition, conventional deep-learning-methods use RNNs [16], [19], [20], [21], [23] to feed a sequence of skeletons as a sequence of vectors, or use CNNs [24], [25], [26], [27] to extract features from a 2D pseudo-image that represents a sequence of skeletons. Recently, a graph convolutional network (GCN)-based method was proposed and drew attention owing to its achievement of high performance. GCN represents joints as vertices and their natural connections in the human body as edges and then calculates convolution based on vertices connected by edges. Therefore, GCN more naturally models the human body than a sequence vector and 2D pseudo-image. Thus, we focus on those methods using GCNs. Spatial temporal graph convolutional networks (ST-GCN) [7] is the first method to use GCN for action recognition with a sequence of skeletons. ST-GCN includes a spatial graph and temporal graph to input a sequence of skeletons directly and extract features from joints on both the intra- and the inter-frames. Subsequently, many GCN methods representing a more appropriate spatial graph on intra-frames have been proposed [8], [9], [10], [11], [12], [13] and the performance has improved dramatically. However, these methods disregard optimization of the temporal graph on the inter-frame. We show an example in Fig.1. Conventional GCN methods connect between vertices corresponding only to the same joint for the temporal dimension (Fig.1, left). This methodology is certain to be effective in extracting feature of the trajectory of the same joints. However, it is too simple to extract the feature of the correlative movement between each joint on the inter-frame.

Our research goal is to optimize both spatial and temporal graphs for a further performance improvement. In this paper, we propose the temporal extension module (TEM) to extend the temporal graph on the inter-frame for skeleton-based action recognition. Our module directly adds edges to not only the same vertex but also neighboring multiple vertices and calculates convolution based on the same multiple vertices on the inter-frame (Fig.1, right). In Fig.1, right, the vertex corresponding to elbow at time $t$ connects joints corresponding to elbow, wrist, and shoulder at time $t-1$ on the inter-frame as an example. The elbow position at time $t$ and the connected joint position at time $t-1$ are correlated because these adjacent joints often move together in cases of action movement such as a "throw." Therefore, it is useful for action recognition to extract the feature from multiple adjacent joints at time $t-1$ as well as the same joint. Our module is simple yet effective in extracting correlated features of multiple adjacent joints that are connected in human body movement. Moreover, our module aids in further performance improvements, combining with other GCN-methods that optimize only the spatial graph. Implementation is easy because this module is independent of graph convolutions on the spatial graph. It has the benefit of reuse of the existing method. We experiment on two large-scale skeleton datasets to evaluate the performance of our module and show that our module is effective and the model with our module achieves state-of-the-art performance compared with previous state-of-the-art methods.

Our contributions of the paper are the following:

- We propose a temporal extension module for extending the temporal graph on the inter-frame. The module is simple yet effective in extracting correlated features of multiple adjacent joints that are connected in human body movement.
- We show the effectiveness of our module in ablation studies. We implement our module to some state-of-the-art models and conduct extensive experiments to evaluate performance. We show that the model with our module outperforms models without our module on two large-scale datasets.
- We conduct several experiments to evaluate performance and show that the model with our module achieves state-of-the-art performance compared with previous state-of-the-art methods.

## II. RELATED WORK

Extensive research has been conducted on more appropriate representations of a sequence of skeletons for skeleton-based action recognition. There are two methods: handcraft-based and deep-learning based methods. In handcraft-based methods, for example, 3D skeletons are represented as a Lie group [14] or vector-valued function [15]. However, the performance of methods using handcrafted features is limited. Currently, data-driven methods using deep-learning are proposed owing to their high representation capacity. In deep-learning based action recognition, there are three methods: RNN-based [16], [19], [20], [21], [23], CNN-based [24], [25], [26], [27], and GCN-based [7], [8], [9], [10], [11], [12], [13], [22], [33], [34].

The RNN-based methods regard a sequence of skeletons as a sequence of vectors and feed a sequence of skeletons directly into several recurrent neural networks. Song et al. [16] introduce an attention module [17] and long short-term memory [18] for the selection of important joints in temporal and spatial dimensions. The CNN-based methods represent a sequence of skeletons as some 2D pseudo-image and extract features through several convolutional neural networks. Li et al. [24] introduce a skeleton transformer module to select important skeleton joints. Liu et al. [26] introduce a method to transform a sequence of skeletons into the 2D pseudo-image, which is robust for view variations. However, these improvements are limited because it disregards the kinematic structure of the human body.

Recently, the GCN-based method has been proposed and has received attention due to its achievement of high performance. The GCN-based method represents a sequence of skeletons as a graph. In the graph, joints correspond to vertices and natural connections of the human body correspond to an edge. This representation has enabled extraction of features based on vertices connected by an edge. Therefore, the GCN-based method can represent the kinematic structure of the human body more naturally than the RNN-based and CNN-based methods. There are two approaches to extract feature from a graph: spectral-based approaches [33], [34] and spatial-based approaches [7], [8], [9], [10], [11], [12], [13], [22]. Spectral-based approaches perform a graph convolution of input signals based on the spectral graph theory. Spatial-based approaches perform convolution directly on the vertices and their neighboring vertices. Since spatial-based approaches do not require high computational complexity, we follow spatial-based approaches in this work. Spatial temporal graph convolutional networks (ST-GCN) [7] is the first method to use a graph convolutional neural network for action recognition with a sequence of skeletons. The spatial graph in ST-GCN is fixed, and the edge is connected between joints in a human natural body only. Therefore, Shi et al. propose a two-stream adaptive graph convolutional network (2s-AGCN) [9] that learns optimal edges in the spatial graph. They also propose a directed graph neural network (DGNN) [10] and a multi-stream adaptive graph convolutional network (MS-AAGCN) [12]. DGNN [10] intro-duces two-stage training processes to learn the optimal edges more flexibly and MS-AAGCN [12] introduces attention modules and multi-stream networks to 2s-AGCN [9]. These GCN-based methods mainly modify the structure of the spatial graph; however, the structure of the temporal graph remains the same as [7]. Gao et al. [33] introduce a graph regression based GCN (GR-GCN) that enables extending a temporal graph and learning of edge weights on the extended graph from training data. However, since graph regression module and GCN are not trained end-to-end learning, the edge weights are not optimum. Gao et al. [22] introduce a latent node to capture temporal contextual information. However, the information on the correlation between each joint on the inter-frame is lost because the features of all joints on the intra-frame are aggregated on only one latent node.

## III. Graph Convolutional Networks

In this section, we show a basic graph construction and the operation of graph convolution on a sequence of skeletons.

### A. Graph Construction on a Sequence of Skeletons

A sequence of skeletons in one sample is represented as a sequence of vectors. Each vector corresponds to spatially and temporally ordered joints of the 2D or 3D position. Yan et al. [7] introduces a spatial-temporal graph to model such a sequence. Therefore, we review the spatial-temporal graph in [7]. As shown in Fig.1, left, the graph represents joints as vertices and natural connections of the human body as edges on the intra-frame. Each vertex is also connected to the vertex corresponding to the same joint for the temporal dimension.

### B. Spatial Graph Convolution

Here, we consider graph convolution above the graph in the single frame case for simplicity. The output of graph convolution $f_{\text{out}}(v_i)$ for $i$-th vertex $v_i$ on the intra-frame can be defined as [7]:

$$f_{\text{out}}(v_i) = \sum_{v_j \in B^S(v_i)} \frac{1}{Z_i(v_j)} f_{in}(v_j) \cdot w\big(l_i(v_j)\big), \quad (1)$$

where $f_{in}(v_j)$ is feature map for $v_j$ and $w$ is the weighting function that calculates a weight vector on given input. $B^S(v_i)$ denotes sampling area for $v_i$ and can be written as [7]:

$$B^S(v_i) = \{v_j | d(v_j, v_i) \leq D^S\}, \quad (2)$$

where $d(v_j, v_i)$ denotes the minimum length of any path from $v_i$ to $v_j$. $D^S$ is the maximum length for sampling on the intra-frame. In this work, $D^S$ is set to 1 as with [7].

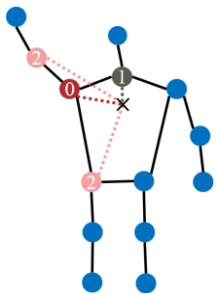

Fig. 2. Subsets for spatial convolution. We draw body joints with dots and joint connection with a solid line. The cross marks the center of gravity. The numbers in dots represent indices of subsets on elbow.

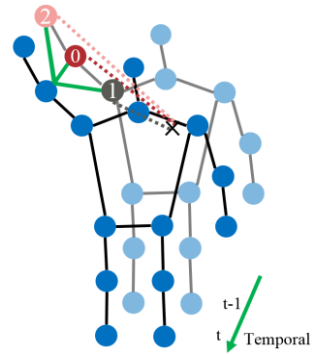

Fig. 3. Subsets for temporal convolution of our module. The numbers in dots represent indices of subsets on elbow at time $t$. We set multiple subsets to represent kinematics correlation as with [7].

In (1), $l_i(v_j)$ denotes the label map on $v_j$ at $v_i$ and empirically divides $B^S(v_i)$ into three kinds of subsets, namely vertex $v_i$ itself, neighboring vertices whose lengths to center of gravity are shorter than $v_i$, and the remainder of the neighboring vertices. Fig.2 show the example of subsets. The normalizing term $Z_i(v_j)$ denotes the cardinality of the corresponding subset.

In implementation, (1) is transformed as [7], [28]:

$$\mathbf{f}_{\text{out}} = \sum_{k}^{K^S} \left( \left( \mathbf{A}_k^S \odot \mathbf{M}_k^S \right) \mathbf{f}_{\text{in}} \right) \mathbf{W}_k^S, \quad (3)$$

where $\mathbf{f}_{\text{out}} \in \mathbb{R}^{N \times C_{\text{out}}}$ denotes output on input feature map $\mathbf{f}_{\text{in}} \in \mathbb{R}^{N \times C_{\text{in}}}$, $N$ is number of joints, $C_{\text{out}}$ is number of output channels and $C_{\text{in}}$ is number of input channels in one frame. In practice, for spatial temporal cases, we can represent the input feature map as the tensor ($N \times F \times C_{\text{in}}$), where $F$ is the number of frames in one sample. $\mathbf{A}_k^S \in \mathbb{R}^{N \times N}$ is the normalized adjacency matrix, i.e., $\mathbf{\Lambda}_k^{-\frac{1}{2}} \overline{\mathbf{A}_k^S} \mathbf{\Lambda}_k^{-\frac{1}{2}}$. $\mathbf{\Lambda}_k \in \mathbb{R}^{N \times N}$ denotes the degree matrix for normalization. $\overline{\mathbf{A}_k^S} \in \mathbb{R}^{N \times N}$ denotes a summation of adjacency matrix $\mathbf{A}_{i,j}^S \in \{0,1\}^{N \times N}$ and identity matrix $\mathbf{I}$ for self-loops. On $\mathbf{A}_{i,j}^S$, $(i,j)$-th $\mathbf{A}_{i,j}^S = 1$ when the $i$-th and the $j$-th joints are connected with bone; otherwise, $\mathbf{A}_{i,j}^S = 0$. $\odot$ is the element-wise product. $\mathbf{M}_k^S \in \mathbb{R}^{N \times N}$ is the learnable matrix for multiplying weight to each vertex. $\mathbf{W}_k^S \in \mathbb{R}^{C_{\text{in}} \times C_{\text{out}}}$ denotes a weight matrix for $1 \times 1$ convolution.

### C. Temporal Graph Convolution

Yan et al. [7] also define graph convolution for temporal dimension. They performed simple $\Gamma \times 1$ convolution to $\mathbf{f}_{\text{out}}$ in (3), where $\Gamma$ denotes kernel size for the temporal dimension. Therefore, sampling area $B^T(v_{ti})$ for the temporal dimension on $v_{ti}$ can be written as [7]:

$$B^T(v_{ti}) = \left\{ v_{qi} \middle| |q - t| \leq \left\lfloor \frac{\Gamma}{2} \right\rfloor \right\}. \quad (4)$$

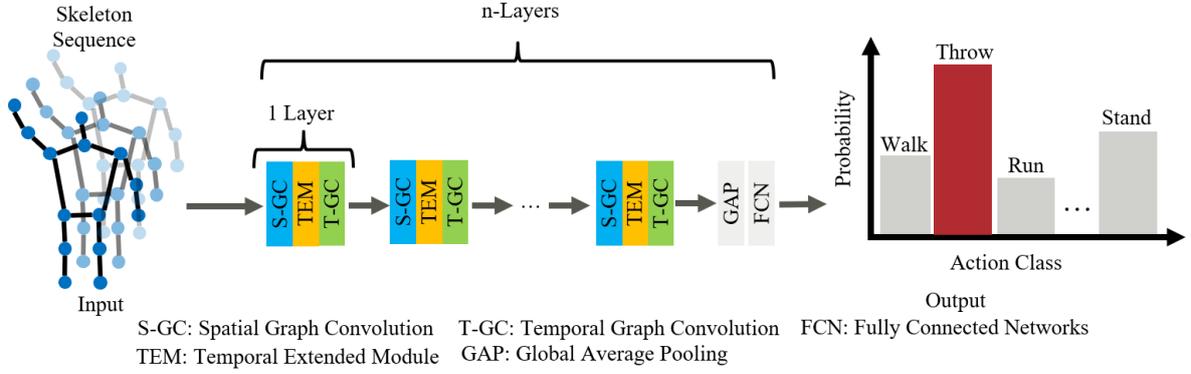

Fig. 4. Implementation of our temporal extension module (TEM) in ST-GCN[7]. ST-GCN[7] consists of multiple layers containing a spatial graph convolution and a temporal graph convolution. We therefore attach one of our modules between these convolutions per a layer.

In this work, $\Gamma$ is set to 9 as with [7].

## IV. TEMPORAL EXTENSION MODULE

In this section, we introduce the temporal extension module (TEM) in detail and the temporal convolution added for the module.

### A. Temporal Extension Module

The conventional temporal graph convolution written in Sec. III calculates based on vertices corresponding to the same joints for temporal dimension. This way does not extract features of the neighboring vertices directly except at the same time. Therefore, it is too simple to extract the feature of the correlative movement between each joint on the inter-frame as explained in Sec. I.

To solve this limitation, we propose the temporal extension module. Our module adds edges to the vertices corresponding not only to the same joint but also to multiple adjacent joints on the inter-frame (Fig.1, right). In general, GCN-based method first extracts the spatial features at a current time and then extracts the temporal features for the long-term [7], [9], [12]. Our module extracts the features on inter-frame. To expand the sampling area for the temporal dimension gradually, we attach our module between conventional spatial graph and temporal graph convolution (i.e., conventional spatial graph convolution is calculated first, then temporal convolution of our module is calculated, and finally, conventional temporal convolution is calculated). We do not change the structure of conventional spatial graph convolution and temporal convolution at all. Therefore, our module can readily apply to many existing methods.

The output of the temporal convolution of our module $f_{out}(v_{ti})$ for $i$-th vertex $v_i$ at time $t$ can be written as

$$f_{out}(v_{ti}) = \sum_{v_{(t-1)j} \in B^T(v_{ti})} \frac{1}{Z_{ti}(v_{(t-1)j})} f_{in}(v_{(t-1)j}) \cdot w\left(l_{(t-1)i}(v_{(t-1)j})\right) \quad (5),$$

where $l_{(t-1)i}(v_{(t-1)j})$ denotes a label map on $v_{(t-1)j}$. (5) is similar to (1). The difference between (5) and (1) is that the convolution on (5) is calculated from vertices at time $t - 1$, while the convolution on (1) is calculated from vertices at time $t$. In other words, our module extract features from multiple joints on the inter-frame. $B^T(v_{ti})$ denotes sampling area on the inter-frame for $v_{ti}$ and can be written as

$$B^T(v_{ti}) = \left\{ v_{(t-1)j} \middle| d(v_{(t-1)j}, v_{(t-1)i}) \leq D^T \right\}, \quad (6)$$

where $D^T$ is maximum length for sampling on the inter-frame. In this work, $D^T$ is set to 1. $B^T(v_{ti})$ is divided into three kinds of subsets such as on conventional spatial convolution. Fig.3 shows the example of subsets on the elbow at time $t$. Comparing the sampling area with the sampling area of the conventional temporal convolution in (4), the temporal convolution of our modules chooses multiple neighboring joints on the inter-frame, while the conventional temporal convolution chooses a certain trajectory of the same joints for the temporal dimension. Therefore, our module helps to extract correlated features of multiple adjacent joints connected in human body movement.

In implementation, (5) is transformed into

$$\mathbf{f}_{out}^t = \sum_k^{K^T} \left( \left( \mathbf{A}_k^T \odot \mathbf{M}_k^T \right) \mathbf{f}_{in}^{t-1} \right) \mathbf{W}_k^T, \quad (7)$$

where $K^T$ is the number of subsets on the inter-frame described above. In this work, we set $K^T$ to 3. $\mathbf{f}_{in}^{t-1} \in \mathbb{R}^{N \times C_{in}}$ equals output of spatial graph convolution $\mathbf{f}_{out}$ at time $t - 1$ in (3). $\mathbf{A}_k^T \in \mathbb{R}^{N \times N}$ is a normalized adjacency matrix such as $\mathbf{A}_k^S$. The difference is that $\mathbf{A}_k^T$ refers to connection of joints on the inter-frame while $\mathbf{A}_k^S$ refers to connection of joints on the intra-frame and is the same as $\mathbf{M}_k^T$ and $\mathbf{W}_k^T$.

## B. Implementation Detail

We show the implementation of our module in ST-GCN [7] as the example in Fig.4. ST-GCN[7] consists of multiple layers for the operation of spatial and temporal graph convolution as described above, including a global average pooling layer, a fully connected network and a *softmax* layer for action recognition. This one layer contains both spatial graph and temporal graph convolution. Therefore, per this one layer, we place one of our modules between these convolutions. Many graph convolutional networks for skeleton-based action recognition contain such multiple layers that include both spatial graph and temporal graph convolution separately [9], [12]. We therefore implement our module readily to such models in the same way.

## V. EXPERIMENTS

To evaluate the performance of our module, we experiment on two large-scale skeleton datasets: NTU RGB+D and Kinetics-skeleton for an action recognition task. We perform ablation studies on these datasets to validate the effectiveness of our module. Next, we compare some models that added our modules with other state-of-the-art methods.

## A. Datasets

### 1) NTU RGB+D

NTU RGB+D [29] is a large-scale and multi-modality dataset for skeleton-based action recognition. It consists of 56,880 video clips in 60 action classes (40 daily actions, nine health-related actions, and 11 mutual actions). These actions are performed by 40 subjects whose ages range from 10 to 35. The action clips are captured by three Microsoft Kinect v2 sensors (sensor 1, sensor 2, and sensor 3). These sensors are located at the same height and three horizontal angles. They set values at -45 degrees for sensor 1, 0 degrees for sensor 2, and 45 degrees for sensor 3. The dataset includes sequences of 3d skeleton that consist of 25 joints, RGB frames, depth maps and IR sequence. We only use sequences of 3d skeleton in this work. Two benchmarks are defined as Cross-Subject (CS) and Cross-View (CV) [29]. In CS, the training sets contain 40,320 samples from 20 subjects and the test sets contain 16,560 samples from the remaining 20 subjects. In CV, the training sets contain 37,920 samples captured by sensor 2 and sensor 3 and the test sets contain 18,960 samples captured by sensor 1. Shahroudy et al. [29] report top-1 accuracy on both benchmarks. Therefore, we follow these protocols.

### 2) Kinetics-Skeleton

Kinetics-Skeleton [7] is a large-scale dataset for skeleton-based action recognition. Original Kinetics [31] contains about 300,000 video clips in 400 classes collected from YouTube. The original Kinetics does not contain joint information, so Yan. et al. [7] estimate the 2D 18 joint coordinates and confidence per person using the OpenPose [5] toolbox. The Open-Pose toolbox is publicly available. The released dataset is divided into training sets (240,000 clips) and test sets (20,000 clips). The clips have 300 frames and each frame contains top-2 persons for high confidence. Yan et al. [7] reports top-1 and top-5 accuracy on the Kinetics-Skeleton. Therefore, we follow the protocol.

## B. Model using Our Module for Experiments

We compare the performance of some models with and without our module. We select the current three state-of-the-art models, ST-GCN [7], 2s-AGCN [9], and MS-AAGCN [12], that are available for official implementation. Next, we implement our module in these models. We call these implemented models ST-GCN+TEM, 2s-AGCN+TEM, and MS-AAGCN+TEM. These models contain multiple layers, including spatial graph and temporal graph convolution for each layer. Therefore, we attach our module between these spatial graph and temporal graph convolution as described in Sec. IV.

## C. Training Detail

All experiments are conducted on the PyTorch deep learning framework [29]. We apply a stochastic gradient descent with Nesterov momentum for optimization strategy. We set Nesterov momentum, weight decay, and initial learning rate to 0.9, 0.0001, 0.1, respectively. We follow same preprocessing and hyper parameter as in ST-GCN [7], 2s-AGCN [9], and MS-AAGCN [12] for fair comparison except batch size. We set batch size of all models to 32 on NTU RGB+D. On Kinetics-Skeleton, we set batch size to 128 on ST-GCN, ST-GCN+TEM, 2s-AGCN, and 2s-AGCN+TEM, and to 64 on MS-AAGCN and MS-AAGCN+TEM. These numbers are maximum values on our GPUs. We report the best result after performing the weighted-score-level fusion on a two-stream (joints and bones) for 2s-AGCN+TEM and a four-stream (joints, bones, joint motion, and bone motion) for MS-AAGCN+TEM. We search weighted parameters for the fusion by a hyperparameter optimization framework, Optuna [32].

## D. Ablation Study

To validate the effectiveness of our module, we compare the performance of ST-GCN and ST-GCN+TEM, 2s-AGCN and 2s-AGCN+TEM, MS-AAGCN and MS-AAGCN+TEM.

We show the evaluation results in Table I. For ST-GCN and ST-GCN+TEM, TEM brings improvements of 2.6% and 1.5% on CS and CV benchmarks on NTU RGB+D, respectively. TEM also brings improvements of 2.0% and 1.8% on Top-1 and Top-5 benchmarks on the Kinetics-skeleton, respectively. For 2s-AGCN and 2s-AGCN+TEM, TEM brings improvements of 0.1% and 0.6% on CS and CV benchmarks, respectively, on NTU RGB+D. TEM also brings improvements of 1.9% and 1.8% on Top-1 and Top-5 benchmarks, respectively on the Kinetics-skeleton. For MS-AAGCN and MS-AAGCN+TEM, TEM brings improvements of 0.7% and 0.4% on CS and CV benchmarks, respectively, on NTU RGB+D.

TEM also brings improvements of 0.6% and 0.8% on Top-1 and Top-5 benchmarks, respectively, on the Kinetics-skeleton. These results show the effectiveness of our module. The results also show that the best performance model is MS-AAGCN+TEM on NTU RGB+D and 2s-AGCN+TEM on the Kinetics-Skeleton.

*E. Comparison with State-of-the-arts Methods*

We compare the best performance model in ablation study with several state-of-the-art methods on NTU RGB+D and Kinetics-skeleton datasets respectively. We select MS-AAGCN+TEM from the NTU RGB+D dataset and 2s-AGCN+TEM from Kinetics-Skeleton for the comparison. The state-of-the-arts methods for comparisons is including RNN-based [16], [19], [20], [23], CNN-based [24], [25], [26], [27], GCN-based [7], [8], [9], [10], [11], [12], [13], [22], [33], [34]. We show the result on Table II and Table III on NTU RGB+D and Kinetics-skeleton datasets respectively. For NTU RGB+D, MS-AAGCN+TEM achieves state-of-the-art performance on both CV and CS. For Kinetics-skeleton, 2s-AGCN+TEM achieves the state-of-the-art performance on Top-1 and Top-5.

## VI. CONCLUSION

In this paper, we propose a TEM for skeleton-based action recognition. The module is simple yet effective for extracting the feature of neighboring multiple joints connected in human body movement by extending the temporal graph on the inter-frame. We conducted extensive experiments on two very large datasets, NTU RGB+D and Kinetics-Skeleton. As a result, we showed the effectiveness of our module and showed that the model in which our module is implemented achieves state-of-the-art performance compared with previous state-of-the-art methods on both of them.

TABLE I. COMPARISONS OF THE RECOGNITION ACCURACY THE MODELS WITH TEM AND WITHOUT TEM

| Methods | NTU-RGB+D | | Kinetics-Skeleton | |
|---|---|---|---|---|
| | *CS (%)* | *CV (%)* | *Top-1 (%)* | *Top-5 (%)* |
| ST-GCN | 82.6 | 88.7 | 32.5 | 54.9 |
| ST-GCN+TEM | 85.2 | 90.2 | 34.5 | 56.7 |
| 2s-AGCN | 88.6 | 95.2 | 36.7 | 59.8 |
| 2s-AGCN+TEM | 88.7 | 95.8 | 38.6 | 61.6 |
| MS-AAGCN | 90.3 | 96.1 | 37.4 | 60.6 |
| MS-AAGCN+TEM | 91.0 | 96.5 | 38.0 | 61.4 |

TABLE II. COMPARISONS OF THE RECOGNITION ACCURACY WITH MS-AAGCN+TEM AND CURRENT STATE-OF-THE-ART METHODS ON NTU RGB+D DATASET

| Methods | NTU-RGB+D | |
|---|---|---|
| | *CS (%)* | *CV (%)* |
| ST-LSTM (Tree Traversal) + Trust Gate [19] | 69.2 | 77.7 |
| TSRJI (Late Fusion) [27] | 73.3 | 80.3 |
| STA-LSTM [16] | 73.4 | 81.2 |
| ESV (Synthesized+Pre-trained) [26] | 80.0 | 87.2 |
| VA-LSTM [20] | 79.6 | 87.6 |
| ST-GCN [7] | 81.5 | 88.3 |
| Si-GCN [34] | 84.2 | 89.1 |
| CNN-based [24] | 83.2 | 89.3 |
| ARRN-LSTM [23] | 81.8 | 89.6 |
| DPRL+GCNN [8] | 83.5 | 89.8 |
| multi-scale network (ResNet152 + 3scale) [25] | 85.0 | 92.3 |
| Complete GR-GCN model [33] | 87.5 | 94,3 |
| 2s-AGCN [9] | 88.5 | 95.1 |
| GCN-NAS [11] | 89.4 | 95.7 |
| DGNN [10] | 89.9 | 96.1 |
| MS-AAGCN [12] | 90.0 | 96.2 |
| BAGCN [22] | 90.3 | 96.3 |
| Sym-GNN [13] | 90.1 | 96.4 |
| MS-AAGCN+TEM(Ours) | **91.0** | **96.5** |

TABLE III. COMPARISONS OF THE RECOGNITION ACCURACY WITH 2s-AGCN+TEM AND CURRENT STATE-OF-THE-ART METHODS ON THE KINETICS-SKELETON

| Methods | Kinetics-Skeleton | |
|---|---|---|
| | *Top-1 (%)* | *Top-5 (%)* |
| ST-GCN [7] | 30.7 | 52.8 |
| 2s-AGCN [9] | 36.1 | 58.7 |
| DGNN [10] | 36.9 | 59.6 |
| GCN-NAS [11] | 37.1 | 60.1 |
| Sym-GNN [13] | 37.2 | 58.1 |
| BAGCN [22] | 37.3 | 60.2 |
| MS-AAGCN [12] | 37.8 | 61.0 |
| 2s-AGCN+TEM (Ours) | **38.6** | **61.6** |